\documentclass[11pt,a4paper]{article}
\usepackage{tablefootnote}
\usepackage[hyperref]{emnlp2020}
\usepackage{times}
\usepackage{latexsym}
\usepackage{amsmath}
\usepackage{enumitem}
\usepackage{float}
\usepackage{pifont}
\usepackage{graphicx}
\usepackage{graphics}
\usepackage[T1]{fontenc}
\usepackage{footmisc}
\usepackage[]{algorithm2e}
\usepackage{fancyvrb}
\usepackage{xcolor}
\usepackage{multirow}
\usepackage{array}
\usepackage{numprint}
\usepackage{enumitem}
\usepackage{cleveref}
\usepackage{booktabs}
\usepackage[normalem]{ulem}
\usepackage{microtype}

\newcommand\roberta{RoBERTa}
\newcommand\aristo{\textsc{Aristo}}
\newcommand\qasc{\textsc{QASC}}
\newcommand\race{\textsc{Race}}

\newcommand\opt{\textsc{$O$}}
\newcommand\ctx{\textsc{$C$}}
\newcommand\que{\textsc{$q$}}

\newcommand{\ignore}[1]{}

\aclfinalcopy 
\def\aclpaperid{2575} 

\title{What do we expect from Multiple-choice QA Systems?}

\author{Krunal Shah \and Nitish Gupta \and Dan Roth \\
  University of Pennsylvania \\
  \texttt{\{shahkr, nitishg, danroth\}@seas.upenn.edu}
 }

\date{}

\begin{document}
\maketitle

\begin{abstract}
The recent success of machine learning systems on various QA datasets could be interpreted as a significant improvement in models' language understanding abilities. However, using various perturbations, multiple recent works have shown that good performance on a dataset might not indicate performance that correlates well with human's expectations from models that ``understand" language. In this work we consider a top performing model on several Multiple Choice Question Answering (MCQA) datasets, and evaluate it against a set of expectations one might have from such a model, using a series of zero-information perturbations of the model's inputs. Our results show that the model clearly falls short of our expectations, and motivates a modified training approach that forces the model to better attend to the inputs. We show that the new training paradigm leads to a model that performs on par with the original model while better satisfying our
expectations.\footnote{Resources for this work are available at: \\ \href{http://cogcomp.org/page/publication_view/913}{http://cogcomp.org/page/publication\_view/913}}
\end{abstract}
\section{Introduction}
\label{sec:intro}
\begin{figure}
\centering
\includegraphics[width=7.6cm,height=3.6cm]{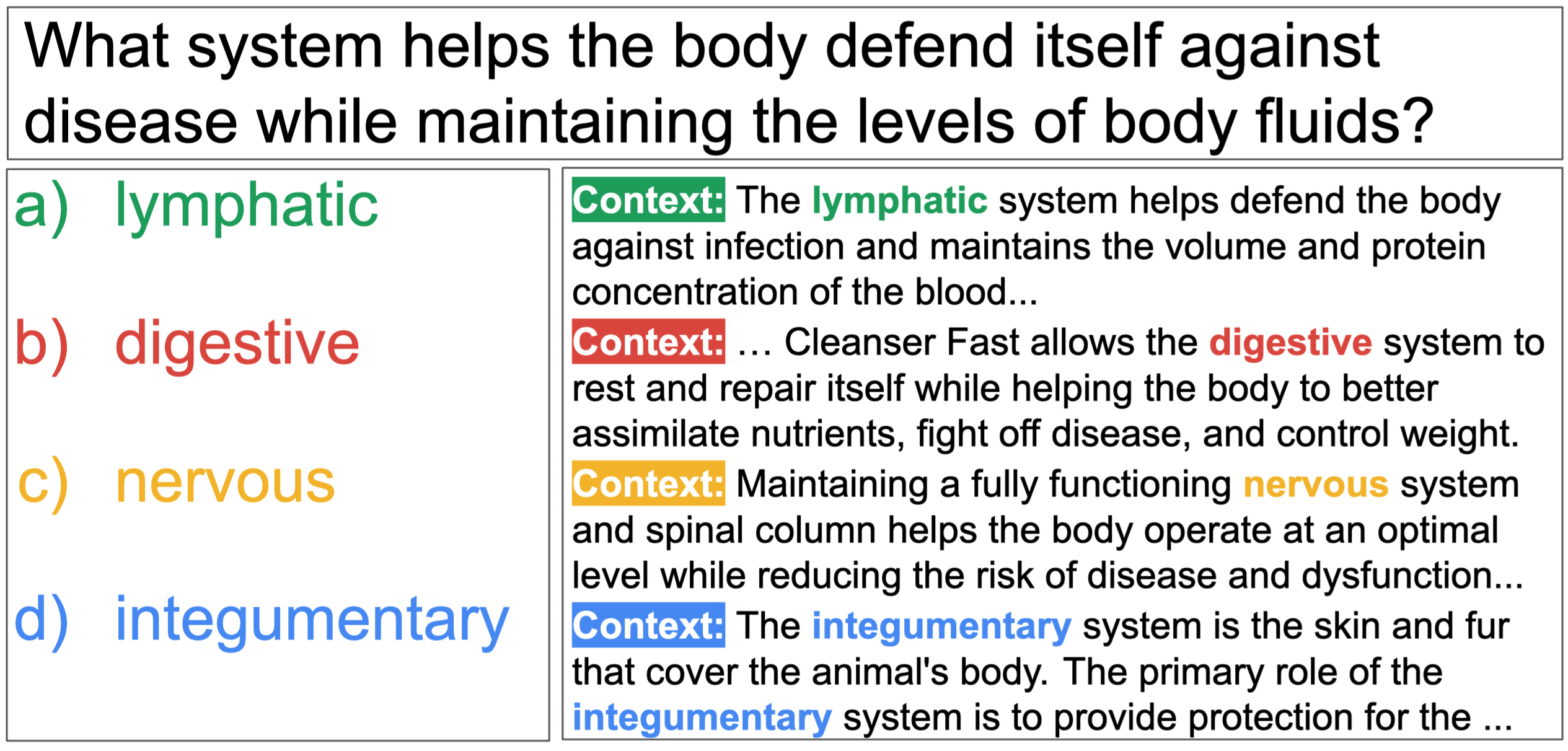}
\caption{\label{example} An example from ARC Easy dataset \cite{clark:2018} showing the three MCQA task inputs. 
}
\end{figure}

Question answering (QA) has been a prevalent format for gauging advances in language understanding. Recent advances in contextual language modelling have led to impressive results on multiple NLP tasks, including on several multiple choice question answering (MCQA, depicted in Fig.~\ref{example}) datasets, a particularly interesting QA task that provides a flexible space of candidate answers along with a simple evaluation.

However, recent work \cite[][inter alia]{KKSCER16, jia-liang-2017-adversarial, si2019does, gardner-etal-2019-making, trivedi-etal-2020-multihop} 
has questioned the interpretation of these QA successes as progress in natural language understanding. Indeed, they exhibit, in various task settings, the brittleness of neural models to various perturbations. They also show ~\cite{kaushik-lipton-2018-much, gururangan-etal-2018-annotation} how models could learn to latch on to spurious correlations in the data to achieve high performance on a given dataset.
In this paper we continue this line of work with a careful analysis of the extent to which the top performing MCQA model satisfies one's expectation from a model that ``understands" language.

We formulate the following set of (non-exhaustive) {\em expectation principles} that a MCQA model should satisfy. 

\phantomsection
\label{robustness}
\noindent \textbf{Monotonicity Expectation:}
Model performance should not drop if an {\em incorrect} option is changed to make it even less likely to be correct.

\phantomsection
\label{behavioral}
\noindent \textbf{Sanity Expectation:} Model should perform poorly given trivially insufficient input. 

\phantomsection
\label{contextual}
\noindent \textbf{Reading Expectation:}
Model should only choose an answer that is  supported by the provided context (and thus perform poorly in the absence of informative context).

While we view the first two expectation principles as necessary axioms, the third could depend on one's definition of the MCQA task. An alternate definition could expect the MCQA model to answer questions using the provided context or, in its absence, using its internal knowledge. In this work, however, we use the Reading Expectation as phrased above; we believe that requiring a model to rely on externally supplied context better gauges its language understanding abilities, and levels the playing field among models with varying levels of internal knowledge.

Guided by these expectation principles we formulate concrete input perturbations to evaluate whether a model satisfies these expectations. 
We show that the top MCQA model fails to meet any of the expectation principles described above. Our results point to the presence of dataset artifacts which the model uses to solve the datasets, rather than the underlying task. 

With goals and insights, we then propose
(a) a different training objective -- which encourages the model to score each candidate option on its own merit, and (b) an unsupervised data augmentation technique -- which aims at ``explaining'' to the model the necessity of simultaneously attending to all inputs, to help the model solve the task.
Our experiments on three popular MCQA datasets indicate that a model trained using our proposed approach better satisfies the expectation principles described above, while performing competitively as compared to the baseline model.
\section{Multi-choice Question Answering}
\label{sec2}
In this section, we briefly describe the multiple-choice question answering (MCQA) task, and the model and datasets we use in this work.\footnote{All results are reported on the dev split of the datasets.}

\paragraph{MCQA Task} 
In a $k$-way MCQA task, a model is provided with a question \que{}, a set of candidate options 
$\mathcal{O} = \{\opt{}_1,\ldots,\opt{}_k\}$, and a supporting context for each option $\mathcal{C} = \{\ctx{}_1, \ldots, \ctx{}_k\}$. The model needs to predict the correct answer option that is best supported by the given contexts. Figure~\ref{example} shows an example of a $4$-way MCQA task.

\paragraph{Datasets} 
We use the following MCQA datasets:
\begin{enumerate}[topsep=3pt,leftmargin=*,itemsep=3pt]
\item
\textbf{\race{}}~\cite{lai-etal-2017-race}: A reading comprehension dataset containing questions from the English exams of middle and high school Chinese students. The context for all options is the same input paragraph.

\item
\textbf{\qasc{}}~\cite{khot2019qasc}: An MCQA dataset containing science questions of elementary and middle school level, which require composition of facts using common-sense reasoning. 

\item
\textbf{\aristo{}}: A collection of non-diagram science questions from standardized tests as used by \citet{clark:2019}.\footnote{Containing questions from the ARC datasets \cite{clark:2018}, NY Regents exams and OBQA\cite{mihaylov-etal-2018-suit}.}

\end{enumerate}
For \qasc{} and \aristo{}, the context for an option is a set of top retrieved sentences as suggested by \citet
{khot2019qasc} and \citet{clark:2019}.

\paragraph{Baseline Model}
We use the RoBERTa large model \cite{liu:2019} for our experiments. Given the task inputs, the model learns to predict a distribution over the candidate options $\mathcal{O}$; which is computed by normalizing the scores for each candidate (using softmax) and the model is trained using cross entropy loss. To compute the score for the $i$-th candidate option $\opt{}_i$, the RoBERTa model is fed with the sequence 
``$\text{[CLS]}\ \ctx{}_i\ \text{[SEP]}\ \que{}\ \text{[SEP]}\ \opt_i\ \text{[SEP]}$'' as input, and the representation of the $\text{[CLS]}$ token is projected to a logit score \cite{clark:2019}.

For \aristo{} and \qasc{}, we first fine-tune the \roberta{} model on \race{}, as suggested by \citealt{clark:2019, khot2019qasc}, and then on the respective datasets.
More details on the training procedure can be found in the appendix.
\section{Model vs. Our Expectations} \label{sec3}
In this section, we define the perturbations we design to evaluate a model against our expectation principles~(defined in \S\ref{sec:intro}). We then analyze how well the baseline model satisfies these expectations.

\paragraph{Monotonicity Expectation:} The following setting tests whether a model is fooled by an obviously incorrect option, one with high word overlap between its inputs.
\begin{itemize}[topsep=0pt,leftmargin=*]
\item \textbf{Perturbed Incorrect Option} (\texttt{\textbf{PIO}}):  The option description for an incorrect option is changed to the question itself and its corresponding context is changed to $10$ concatenations of the question.\footnote{To approximately simulate a typical context's length.} 
\end{itemize}

\paragraph{Sanity Expectation:} The following settings test how the model's performance changes when given an unreasonable input, for which it should not be possible to predict the correct answer. 
\begin{itemize}[topsep=0pt,leftmargin=*,itemsep=0pt]
\item \textbf{No option} (\texttt{\textbf{NO}}): The option descriptions for all candidate options is changed to empty, ``<s>".
\item \textbf{No question} (\texttt{\textbf{NQ}}): The question (for all its candidate options $\mathcal{O}$) is changed to empty , ``<s>".
\end{itemize}

\paragraph{Reading Expectation:} The following setting tests how crucial the context is for the model to correctly answer the questions.
\begin{itemize}[topsep=0pt,leftmargin=*]
\item \textbf{No context} (\texttt{\textbf{NC}}): The contexts for all candidate options is changed to empty, ``<s>".
\end{itemize}

\paragraph{Baseline model performance}
Table~\ref{table:understand} shows that the model achieves impressive accuracy on all three dataset; \race{} ($84.8$), \qasc{} ($85.2$), and \aristo{} ($78.3$), which suggests that the model should satisfy the expectations laid out for a good MCQA model.

\paragraph{Evaluating expectations}
When evaluating the model by modifying an incorrect option and its context (\texttt{\textbf{PIO}}), we find that its performance drops notably across all three datasets, for example, from $85.2 \rightarrow 7.9$ for \qasc{}. This shows that the model is not able to cope with an incorrect option containing high word overlap with the question, even when it is trivially wrong and the correct option and its context are present. The baseline model thus fails to satisfy the Monotonicity Expectation.

Given an unreasonable input, where a pivotal component of the input is missing, we find that the baseline model still performs surprisingly well. For example, in \aristo{}, removal of the question (\texttt{\textbf{NQ}}) only leads to a performance drop from $78.3 \rightarrow 55.3$, and removal of the options (\texttt{\textbf{NO}}), from $78.3 \rightarrow 46.8$. This suggests that the datasets contain unwanted biases that the model relies on to answer correctly. This shows that the baseline model fails to satisfy the Sanity Expectation.

The model achieves reasonable performance on the removal of the contexts; thus failing our Reading Expectation, e.g., performance only drops from $78.3 \rightarrow 63.8$ in \aristo{}.
To achieve this performance the model must rely on its inherent knowledge \cite{petroni-etal-2019-language} or, more likely, on dataset artifacts as suggested previously.

\begin{table}
\begin{tabular}{lccc}
\toprule
\textbf{Eval. Setting} & \textbf{\aristo{}} & \textbf{\race{}} & \textbf{\qasc{}} \\
\midrule
\textbf{Original ($\uparrow$)}  					& 78.3           & 84.8            & 85.2 \\ 
\midrule
\textbf{Perturbed Inco-}      & \multirow{2}{*}{25.4}           & \multirow{2}{*}{45.8}            & \multirow{2}{*}{7.9} \\
\textbf{rrect Option ($\uparrow$)}    &            &             & \\
\midrule
\textbf{No Option ($\downarrow$)}  	& 46.8           & $-$              & 50.2 \\
\textbf{No Question ($\downarrow$)}    & 55.3           & 62.8            & 34.3 	 \\
\textbf{No Context ($\downarrow$)}    	& 63.8           & 49.1            & 55.8 \\
\bottomrule
\end{tabular}
\caption{\label{table:understand} Accuracy of the respective RoBERTa models on \race{}, \aristo{} and \qasc{} datasets for the different evaluation settings detailed in Section \ref{sec3}. 
The No Option setting is not applicable for \race{} as all options would have the same inputs. The arrows denote the expected performance where $\uparrow$ denotes higher is better and $\downarrow$ denotes that lower performance is better.}
\end{table}
\begin{table*}
\centering
\scalebox{0.98}{
\begin{tabular}{llcccccc}
\toprule
\multirow{3}{*}{\textbf{Dataset}}   & \multirow{3}{*}{\textbf{Model}}  & \multirow{2}{*}{\textbf{Original}} & \textbf{Perturbed} & \textbf{No} & \textbf{No} & \textbf{No}  \\
   &  &  & \textbf{Incorrect Option} & \textbf{Option} & \textbf{Question} & \textbf{Context}\\
  &  & (\texttt{\textbf{O})($\uparrow$)} & (\texttt{\textbf{PIO})($\uparrow$)} & (\texttt{\textbf{NO})($\downarrow$)} & (\texttt{\textbf{NQ})($\downarrow$)} & (\texttt{\textbf{NC})($\downarrow$)} \\ \midrule
\multirow{2}{*}{\textbf{\aristo{}}}    &  RoBERTa                & \textbf{78.3}  & 25.4           & 46.8               & 55.3               & 63.8          \\ 
                                       &  \,+ Our Training           & 75.8           & \textbf{55.5}  & \textbf{26.9}      & \textbf{35.4}      & \textbf{42.4} \\ \midrule
\multirow{2}{*}{\textbf{\race{}}}      &  RoBERTa                & \textbf{84.8}           & 45.8           & $-$                & 62.8               & 49.1           \\ 
                                       &  \,+ Our Training           & 83.9  & \textbf{72.4}  & $-$                & \textbf{12.4}      & \textbf{20.6}  \\ \midrule
\multirow{2}{*}{\textbf{\qasc{}}}      &  RoBERTa                & \textbf{85.2}  & 7.9           & 50.2               & 34.3               & 55.8            \\ 
                                       &  \,+ Our Training           & 82.6           & \textbf{38.0}  & \textbf{13.7}      & \textbf{12.3}      & \textbf{34.7}  \\
\bottomrule
\end{tabular}
}
\caption{Comparison of a model trained using our proposed training approach with the baseline model on \race{}, \aristo{} and \qasc{} datasets. The evaluation settings used are described in Section \ref{sec3}.
\label{table:contrast}} 
\end{table*}

\section{Proposed Training Approach}
To address the aforementioned limitations, and reduce the tendency of the model to exploit dataset artifacts, we propose the following modifications to the training methodology.

\subsection{MCQA as Binary Classification}
Treating MCQA as a multi-class classification problem requires the model to {\em minimally} differentiate the correct option from the incorrect options, thus making the training sensitive to the relative difficulty between the options. 
We propose to prevent this by training the model to predict the \emph{correctness} of each candidate option separately, by converting the $k$-way MCQA task into $k$ binary classification tasks. The model is trained to predict a high probability for the correct option triplet $(q, O_g, C_g)$, and low for the other $k-1$ options.

\subsection{Unsupervised data augmentation}
We introduce an unsupervised data augmentation technique to discourage the model from exploiting spurious correlations between pairs of inputs and encourage it to read \emph{all} the inputs. 
During training, given an MCQA instance $(q, \mathcal{O}, \mathcal{C})$, for each of the option triplet $(q, O_{i}, C_{i})$, we generate new examples (each with negative label) by performing one of the following perturbations with a certain probability (details in the appendix):

\noindent \textbf{Option:} $O_i$ is changed to one of (a) empty (``<s>") or (b) $O_j \in \mathcal{O}; j \neq g$.

\noindent \textbf{Context:} $C_i$ is changed to one of (a) empty (``<s>") or (b) $C_j \in \mathcal{C}; j \neq g$.

\noindent \textbf{Question:} $q$, for all options $\mathcal{O}$ is changed to one of (a) empty (``<s>") or (b) another question from the training set.

\noindent \textbf{No change:} The triple is left as is.

This is an automatic data augmentation and requires no manual annotation.
\section{Results} \label{sec:results}
The performance of
our proposed training approach (+ Our Training) along with  the baseline model are presented in Table~\ref{table:contrast}.
The new model performs competitively (within $2.6$ points) with the baseline on all three datasets suggesting that our proposed training approach only has minor impact on the overall model performance.

In our \texttt{\textbf{PIO}} setting, the new model outperforms the baseline on all three datasets by a large margin ($55.5$ compared to baseline's $25.4$ on \aristo{} indicating an improvement over the baseline with regard to our Monotonicity Expectation. Even though the data augmentation did not augment examples with this perturbation, our training approach helps the model better read the inputs and avoid distractor options.

When evaluating over unreasonable inputs in the \texttt{\textbf{NO}} and \texttt{\textbf{NQ}} settings, the resulting model performs poorly compared to the baseline ($13.7$ vs. $50.2$ and $12.3$ vs. $34.3$ on \qasc{}), showing that our training approach helps the model to not rely on dataset bias and satisfy the Sanity Expectation.

Finally, the new model performs poorly when we remove the contexts (e.g $20.6$ on \race{}), indicating how it is able to meet our Reading Expectations. The results also show the resulting model's reliance on the context for information required to correctly answer questions.
Moreover, it implies that the resulting model is able to achieve performance similar to the baseline by heavily relying on information from the contexts, as opposed to the baseline that exploits dataset artifacts (as previously shown).

Results showing the performance of the model trained using binary classification loss, without the data augmentation, are attached in the appendix.
\section{Related work}
Our work builds on numerous recent works that challenge the robustness of neural language models \cite{jin:2019, si2019does} or, more generally, neural models \cite{kaushik-lipton-2018-much, jia-liang-2017-adversarial, KKSCER16}.
Our evaluation settings -- hiding one of the three inputs to the MCQA models -- are similar to \citealt{kaushik-lipton-2018-much}'s partial input settings which were designed to point out the existence of dataset artifacts in reading comprehension datasets. However, we argue that our results additionally point to a need for more robust training methodologies and propose an improved training approach.
Our data augmentation approach builds on recent works \cite{khashabi2020natural, kobayashi-2018-contextual, kaushik2019learning, cheng-etal-2018-towards, andreas-2020-good} that try to leverage augmenting training data to improve the performance and/or robustness of models. However most of these works are semi-automatic or require human annotation while our augmentation approach requires no additional annotation.
\section{Conclusion}
We formulated three \emph{expectation principles} that a  MCQA model must satisfy, and devised appropriate settings to evaluate a model against these principles. Our evaluations on a \roberta{}-based model showed that the model fails to satisfy any of our expectations, and exposed its brittleness and reliance on dataset artifacts. 
To improve learning, we proposed a modified training objective to reduce the model's sensitivity to the relative difficulty of candidate options, and an unsupervised data augmentation technique to encourage the model to rely on \emph{all} the input components of a MCQA problem.
The evaluation of our proposed training approach showed that the resulting model performs competitively with the original model while being robust to perturbations; hence, closer to satisfying our \emph{expectation principles}.
\section*{Acknowledgements}
We would like to thank Peter Clark, Oyvind Tafjord and Tushar Khot for providing us with their dataset and model configurations.
This work was funded by ONR Contract N00014-19-1-2620, by
the Oﬃce of the Director of National Intelligence (ODNI), Intelligence Advanced Research Projects Activity (IARPA), via IARPA Contract No. 2019-19051600006 under the BETTER Program, and by contract FA8750-19-2-0201 with the US Defense Advanced Research Projects Agency (DARPA). The views expressed are those of the authors and do not reflect the official policy or position of the Department of Defense or the U.S. Government.
 
\bibliography{emnlp2020}
\bibliographystyle{acl_natbib}

\appendix
\section{Training and Evaluation Details} \label{ap_sec3}
We note the following points about the details of our experiments:
\begin{enumerate}
    \item For our experiments on the modified training approach, only the final dataset specific finetuning step is modified to our approach.
    \item For the proposed training method on RACE dataset, the augmented training approach was only applied after the first epoch of training.
    \item Since our augmentation approach only modifies examples on the go, the number of examples the model sees in a single epoch remains the same. The baseline models were all finetuned for $4$ epochs and the models using our proposed approach were finetuned for $1$ extra epoch in all cases.
    \item Training on NVIDIA TITAN RTX GPUs with 24GB of memory, one epoch on \race{}, \aristo{} and \qasc{} required $4$ hours, $40$ minutes and $55$ minutes respectively.
\end{enumerate}
\section{Data augmentation steps}
During training (or prior to it), each example would be modified using the following steps:

\begin{small}
\begin{Verbatim}[commandchars=\\\{\}]
For option opt in all options:
  If isCorrect(opt) and prob(0.2):
      Flip label of option opt
      With equal probability:
         1. With equal probability:
            context = "<s>"
            context = incorrect context
         2. With equal probability:
            option = "<s>"
            option = incorrect option
         3. With equal probability:
            question = "<s>"
            question = previous question
  If isIncorrect(opt) and prob(0.8)
      With equal probability:
         1. With equal probability:
            context = "<s>"
            context = incorrect context
         2. With equal probability:
            option = "<s>"
            option = incorrect option
\end{Verbatim}
\end{small}

\begin{table*}
\centering
\begin{tabular}{llcccccc}
\toprule
\multirow{3}{*}{\textbf{Dataset}}   & \multirow{3}{*}{\textbf{Model}}  & \multirow{2}{*}{\textbf{Original}} & \textbf{Perturbed} & \textbf{No} & \textbf{No} & \textbf{No}  \\
   &  &  & \textbf{Incorrect Option} & \textbf{Option} & \textbf{Question} & \textbf{Context}\\
  &  & (\texttt{\textbf{O})($\uparrow$)} & (\texttt{\textbf{PIO})($\uparrow$)} & (\texttt{\textbf{NO})($\downarrow$)} & (\texttt{\textbf{NQ})($\downarrow$)} & (\texttt{\textbf{NC})($\downarrow$)} \\ \midrule
\multirow{2}{*}{\textbf{\aristo{}}}    &  RoBERTa                & \textbf{78.3}  & 25.4           & 46.8               & 55.3               & 63.8          \\ 
                                       &  + Binary Classification & 75.5           & 41.6       & 39.7  & 51.8       & 61.3       \\ 
                                       &  + Our Training           & 75.8           & \textbf{55.5}  & \textbf{26.9}      & \textbf{35.4}      & \textbf{42.4} \\ \midrule
\multirow{2}{*}{\textbf{\race{}}}      &  RoBERTa                & \textbf{84.8}           & 45.8           & $-$                & 62.8               & 49.1           \\ 
                                       &  + Binary Classification & 83.9           & \textbf{75.1}       & $-$  & 60.0       & 49.4       \\ 
                                       &  + Our Training           & 83.9  & 72.4  & $-$                & \textbf{12.4}      & \textbf{20.6}  \\ \midrule
\multirow{2}{*}{\textbf{\qasc{}}}      &  RoBERTa                & \textbf{85.2}  & 7.9           & 50.2               & 34.3               & 55.8            \\ 
                                       &  + Binary Classification & 84.1           & 11.1       & 45.8  & 39.2       & 54.6       \\ 
                                       &  + Our Training           & 82.6           & \textbf{38.0}  & \textbf{13.7}      & \textbf{12.3}      & \textbf{34.7}  \\
\bottomrule
\end{tabular}
\caption{Results contrasting the performance of the baseline model trained using binary classification loss to the baseline model and the model trained using our proposed training approach on \race{}, \aristo{} and \qasc{} datasets. The evaluation settings used are described in the paper.
\label{abalation}}
\end{table*}

\section{Additional Results}
Results showing the performance of the baseline model trained using binary classification loss are described in Table \ref{abalation}.

\end{document}